\documentclass[times,preprint,10pt]{elsarticle}

\usepackage{amssymb}
\usepackage{amsmath}
\usepackage{graphicx}
\usepackage{booktabs}
\usepackage[table]{xcolor}
\definecolor{top1}{gray}{0.70}
\definecolor{top2}{gray}{0.80}
\definecolor{top3}{gray}{0.90}
\usepackage{caption}
\usepackage{pifont}
\usepackage{float}
\usepackage{makecell}
\usepackage{hyperref}
\usepackage{makecell}
\usepackage{adjustbox}
\journal{Preprint}

\begin{document}

\begin{frontmatter}

\title{FastPano3D: Feed-Forward Indoor Panoramic 3D Reconstruction from a Single Image}

\author[inst1]{Jianqiang Li}
% \ead{24211060915@stumail.xsyu.edu.cn}

\author[inst1]{Liumei Zhang\corref{cor1}}
% \ead{zhangliumei@xsyu.edu.cn}

\author[inst1]{Wenjia Guo}
% \ead{25212060940@stumail.xsyu.edu.cn}

\author[inst2]{Tianlong Feng}
% \ead{tlfeng@stu.xidian.edu.cn}

\author[inst2]{Yongzhi Liao}
% \ead{liaoyongzhi1010@stu.xidian.edu.cn}

\author[inst2]{Di Lu}
% \ead{dlu@xidian.edu.cn}

\author[inst3]{Hanchi Ren}
% \ead{hanchiren@swansea.ac.uk}

\author[inst4]{Jingjing Deng\corref{cor1}}
% \ead{jingjingdeng@bristol.ac.uk}

% =====================================================
% Corresponding Author
% =====================================================

\cortext[cor1]{Corresponding author}

% =====================================================
% Affiliations
% =====================================================

\affiliation[inst1]{
organization={Xi'an Shiyou University},
city={Shaanxi},
country={China}
}
\affiliation[inst2]{
organization={Xidian University},
city={Shaanxi},
country={China}
}
\affiliation[inst3]{
organization={Swansea University},
city={Swansea},
country={United Kingdom}
}
\affiliation[inst4]{
organization={University of Bristol},
city={Bristol},
country={United Kingdom}
}

\begin{abstract}
Recent advances in 3D scene reconstruction have highlighted the intricate trade-offs among rendering quality, inference efficiency, and data dependency. To address the challenge of rapidly reconstructing detailed 3D indoor scenes from minimal input, we introduce \emph{FastPano3D}, an end-to-end framework that directly generates renderable 3D Gaussian representations from a single panoramic image. Unlike perspective-based methods, panoramic images inherently suffer from equirectangular projection distortions and spatially non-uniform feature distributions, making direct feed-forward Gaussian generation particularly challenging. In contrast to existing Gaussian Splatting based methods that rely on multi-view supervision or per-scene optimization, \emph{FastPano3D} employs a lightweight feature encoder, adaptive Gaussian sampling, and a point-cloud-guided refinement strategy to achieve efficient and accurate scene generation without any test-time optimization. Our approach reconstructs high-fidelity 3D scenes within seconds, achieving up to \textbf{156× faster} inference than prior state-of-the-art methods such as Pano2Room, while using only half the parameters. Extensive experiments demonstrate that FastPano3D delivers rendering quality comparable to NeRF- and 3DGS-based reconstructions, establishing a new benchmark for rapid, single-view 3D scene inference.
\end{abstract}

% \begin{highlights}
% \item Proposing FastPano3D, an Equirectangular Projection aware feed-forward framework for single-panorama 3D Gaussian reconstruction, achieving up to 156× speed-up over optimization-based methods.
% \item Proposing an adaptive Gaussian sampling strategy that allocates Gaussians according to scene complexity and texture density, effectively balancing sparsity and redundancy.
% \item Proposing a point-cloud guidance module that extracts keypoints from texture and edge cues to constrain Gaussian distributions and preserve geometric consistency.
% \end{highlights}

\begin{keyword}
Single panoramic image \sep Fast reconstruction \sep Indoor scene \sep 3D Gaussian Splatting \sep Feed-forward inference
\end{keyword}

\end{frontmatter}

%%------------------------------------------------------------
\section{Introduction}
\label{sec:intro}

3D scene reconstruction serves as a cornerstone for a wide range of visual computing applications, including augmented and virtual reality, digital twins, autonomous driving, and robotic vision. With the growing accessibility of consumer-grade panoramic cameras, equirectangular projection (ERP) images have become increasingly popular, as they capture a complete 360° scene from a single viewpoint in one shot. However, reconstructing high-quality 3D scenes from panoramic inputs remains challenging. Most existing approaches depend on multi-view images and employ neural radiance fields (NeRF)~\cite{Mildenhall2020nerf} or 3D Gaussian Splatting (3DGS)~\cite{Kerbl20233dgs} as scene representations, which require per-scene optimization and lead to high computational costs and slow reconstruction speeds. Although 3DGS supports real-time rendering, it still requires a time-consuming optimization phase that relies on dense multi-view data, hindering practical deployment in time-sensitive applications.

Existing panoramic 3D reconstruction methods can broadly be grouped into three categories. Optimization-based NeRF and 3DGS methods~\cite{Malarz2024} achieve high fidelity but require minutes to hours of per-scene optimization. Layout-based approaches can recover coarse room structures efficiently but fail to capture fine geometric details. View-synthesis-based methods~\cite{Guo2024pano2room} combine inpainting and generative models to fill unseen regions, but suffer from spatial inconsistencies and depend heavily on the quality of the underlying generative model. \textbf{None of these approaches support feed-forward inference: they all require per-scene fitting or multi-step generation pipelines, making real-time or offline deployment difficult.}

In parallel, feed-forward Gaussian reconstruction has shown great promise for perspective images. Methods such as Splatter Image~\cite{szymanowicz2024splatter} and pixelSplat~\cite{charatan2024pixelsplat} directly regress per-pixel Gaussian parameters from one or two images via encoder-decoder networks, enabling reconstruction in seconds without any optimization loop. \textbf{However, extending this paradigm to panoramic ERP inputs is non-trivial:} ERP images suffer from severe polar distortion, low-texture regions (e.g., ceilings and walls) are prone to depth estimation errors, and the wide field of view demands different sampling and rendering strategies compared to perspective cameras.

\begin{figure}[t]
  \centering
  \includegraphics[width=\linewidth]{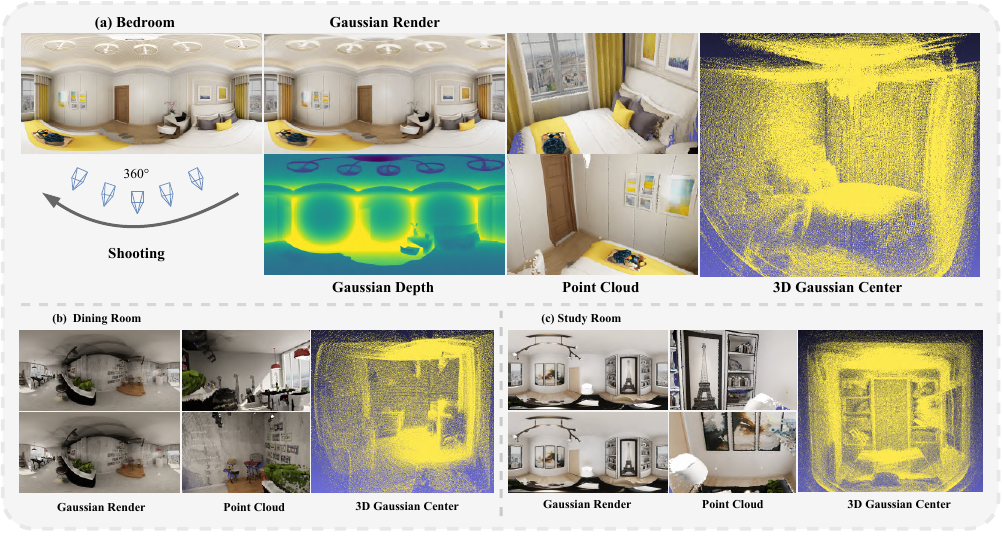}
  \caption{
    In this paper, we present \textbf{FastPano3D}, an ultra-fast end-to-end generative model for Gaussian Splatting that can reconstruct high-fidelity scenes from a single panoramic image in just a few seconds (achieving up to \textbf{156$\times$ speed-up}). In the figure, we showcase the qualitative performance using various indoor scenes, such as (a) ``Bedroom'', (b) ``Dining Room'' and (c) ``Study Room''.
  }
  \label{fig:overview}
\end{figure}

To address these challenges, we propose \emph{FastPano3D}, an end-to-end feed-forward Gaussian reconstruction framework specifically designed for single panoramic images, as illustrated in Fig.~\ref{fig:overview}. \emph{FastPano3D} integrates adaptive Gaussian generation, point-cloud guidance, and cubemap-based rendering into a unified and lightweight pipeline. The framework requires no per-scene optimization: given a single ERP image, it produces a complete 3D Gaussian scene representation in seconds. Our main contributions are as follows:

\begin{itemize}
  \item We present an \textbf{ERP-aware feed-forward Gaussian} reconstruction framework that eliminates per-scene optimization and achieves single-shot panoramic scene inference within seconds, offering up to \textbf{156×} speed-up over optimization-based methods.
  \item We design an \textbf{adaptive Gaussian sampling} strategy that allocates Gaussians according to scene complexity and texture density, effectively balancing sparsity and redundancy.
  \item We introduce a \textbf{point-cloud guidance} module that extracts keypoints from texture and edge cues to constrain Gaussian distributions and preserve geometric consistency.
\end{itemize}

%%------------------------------------------------------------
\section{Related Work}
\label{sec:related}

\subsection{3D Reconstruction from A Single Image}

Single-image 3D reconstruction has made remarkable progress by exploiting geometric cues from 2D images through multi-view prediction, end-to-end differentiable rendering, and NeRF-inspired adaptations. Multi-view CNN~\cite{Tatarchenko2016mutiview} employs convolutional networks to predict RGB and depth maps from arbitrary viewpoints, which are then fused into point clouds and surface meshes for end-to-end 3D inference. Pix2Vox~\cite{Xie2019pix2vox} introduces a context-aware fusion framework that constructs a coarse 3D volume and refines high-quality regions to enhance geometric and texture fidelity. SynSin~\cite{Wiles2020synsin} avoids ground-truth 3D supervision by projecting latent 3D features via a differentiable point cloud renderer and refining occluded or missing regions for realistic novel-view synthesis.

Among depth estimation methods, MiDaS~\cite{Ranftl2022} improves prediction accuracy through training on mixed datasets, DPT~\cite{Ranftl2021} leverages Vision Transformers for precise dense estimation, and MonoDepth introduces self-supervised~\cite{Godard2017} and unsupervised~\cite{Godard2019} constraints to reduce the need for ground-truth depth annotations. More recently, Splatter Image~\cite{szymanowicz2024splatter} proposes a U-Net based architecture to directly regress per-pixel Gaussian parameters from a single perspective image, enabling ultra-fast reconstruction without per-scene optimization. pixelSplat~\cite{charatan2024pixelsplat} extends this paradigm to image pairs, improving geometric consistency and generalization across scenes. More recently, Chen~et~al.~\cite{Chen2023structnerf} address novel view synthesis for indoor scenes with sparse NeRF inputs by exploiting structural hints embedded in multi-view images, while Zhao~et~al.~\cite{Zhao2025generalizable3dgs} propose a generalizable 3DGS framework that bridges 2D image features with 3D Gaussian representations for feed-forward novel view synthesis of unseen scenes. These feed-forward Gaussian methods demonstrate that direct parameter regression can drastically reduce inference time compared to optimization-based 3DGS pipelines, motivating our approach for the panoramic setting.

\subsection{Scene Reconstruction from Panoramic Images}

Panoramic scene reconstruction aims to model complete 3D environments with a full 360° field of view. Early methods approach the problem from a layout estimation perspective: Pano2CAD~\cite{xu2016pano2cad} estimates room geometry and object poses from a single panorama, LayoutNet~\cite{Zou2018layout} and HorizonNet~\cite{Sun2019horizon} recover coarse structural room layouts, and PanoContext~\cite{Zhang2014panocontext} proposes a whole-room context model for panoramic scene understanding. 

For radiance-field-based reconstruction, BiFuse~\cite{BiFuse20} addresses viewpoint shifts through bidirectional projection fusion for panoramic depth estimation. OmniDepth~\cite{zioulis2018omnidepth} proposes an omnidirectional NeRF for panorama-to-3D mapping. PERF~\cite{wang2023PERF} reconstructs visible geometry via volume rendering and fills unseen regions through progressive inpainting guided by diffusion models. PanoNeRF~\cite{Lu2024PanoNeRF} optimizes NeRFs under spherical projections for globally consistent reconstruction, while PanoGRF~\cite{Sun2023PanoGRF} represents scene radiance with Gaussian distributions for fine-grained reconstruction from a single panoramic image.

Optimization-based 3DGS methods such as ODGS~\cite{lee2024odgs} and OmniGS~\cite{li2024omnigs} adapt Gaussian Splatting to omnidirectional inputs, and PanSplat~\cite{Zhang2024PanSplat} targets high-resolution spherical inputs using Gaussian pyramids. Text-driven approaches including FastScene~\cite{ma2024fastscene} and SceneDreamer360~\cite{li2024scenedreamer360} generate indoor scenes from prompts via multi-view inpainting and 3DGS-based modeling. Scene4U~\cite{huang2025scene4u} adopts a hierarchical approach for large-scale outdoor reconstruction. A concurrent line of work targets feed-forward Gaussian splatting for panoramic novel-view synthesis. PanSplat proposes a spherical 3D Gaussian pyramid with a Fibonacci lattice arrangement and achieves feed-forward inference at up to 4K resolution. However, PanSplat requires two wide-baseline panoramas as input and relies on a hierarchical spherical cost volume to establish cross-view correspondences, making it fundamentally a multi-view synthesis method. In contrast, FastPano3D addresses the strictly harder setting of reconstructing a complete 3D scene from a single panoramic image, where no cross-view stereo cues are available. This distinction makes direct quantitative comparison inappropriate, as the two methods operate under different input assumptions and serve complementary use cases.

Despite this progress, all existing panoramic reconstruction methods rely on per-scene optimization or multi-step generation pipelines. A feed-forward framework that directly maps a single ERP image to a complete 3D Gaussian representation without any test-time fitting has not been explored. Our work addresses this gap.

%%------------------------------------------------------------
\section{Method}
\label{sec:method}

\begin{figure}[t]
    \centering
    \includegraphics[width=\linewidth]{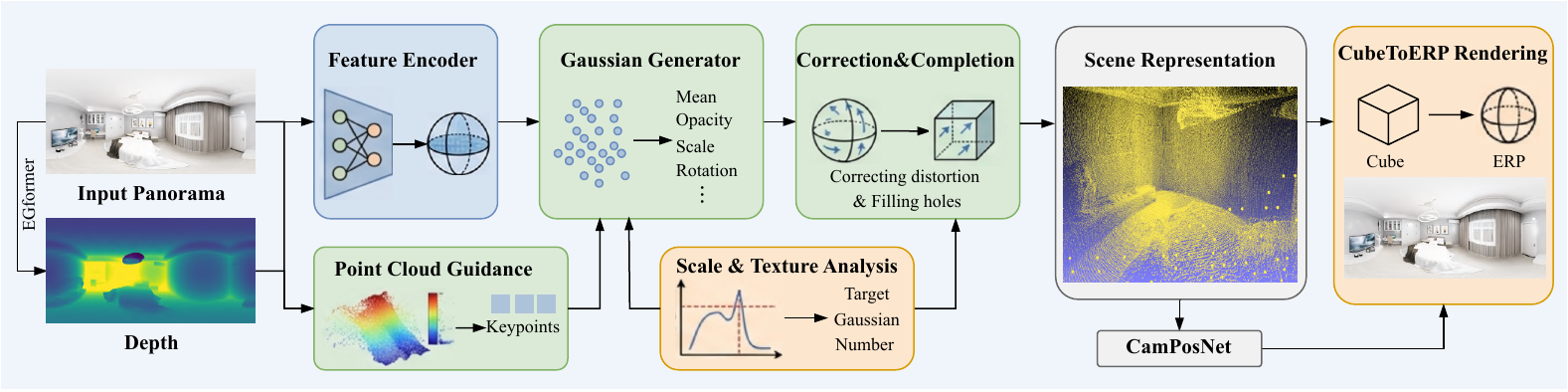}
    \caption{\textbf{Overview of FastPano3D.} Given a single panoramic image, FastPano3D first employs EGformer to predict a dense depth map, which is then lifted into a point cloud to extract geometric keypoints as guidance. A lightweight Feature Encoder extracts multi-scale features from the panorama, which are decoded by the Gaussian Generator into per-Gaussian attributes. The Scale \& Texture Analysis module estimates the target Gaussian number for adaptive density control. The Correction \& Completion module rectifies ERP distortions and inpaints occluded regions using both visual features and point cloud guidance. CamPosNet predicts the camera pose, and the final scene Gaussian representation is rendered back to 2D via CubeToERP, supporting both cubemap and equirectangular outputs, without any per-scene optimization.}
    \label{fig:pipeline}
\end{figure}

\textbf{FastPano3D} takes a single indoor panoramic image as input and produces a full 3D Gaussian scene representation without any per-scene optimization. The pipeline consists of three stages: (1) depth estimation using a pre-trained EGFormer~\cite{Yun2023egformer}, (2) camera scale prediction via a dedicated scale predictor (CamPosNet), and (3) feed-forward Gaussian generation via an adaptive encoder-decoder architecture. The overall pipeline is illustrated in Fig.~\ref{fig:pipeline}.

\subsection{Camera Scale Estimation}

Since the depth predicted by EGFormer is constrained to a relative scale, the training loss requires aligning the predicted scene with ground-truth novel views, which in turn requires a metric scale estimate. Note that because reconstruction is performed in a canonical camera frame, camera rotation and intrinsics do not need to be estimated. The key unknown is the global scene scale, which governs the physical size of the reconstructed geometry and is needed to correctly place training-time supervision 
viewpoints.

\begin{figure}[t]
    \centering
    \includegraphics[width=\linewidth]{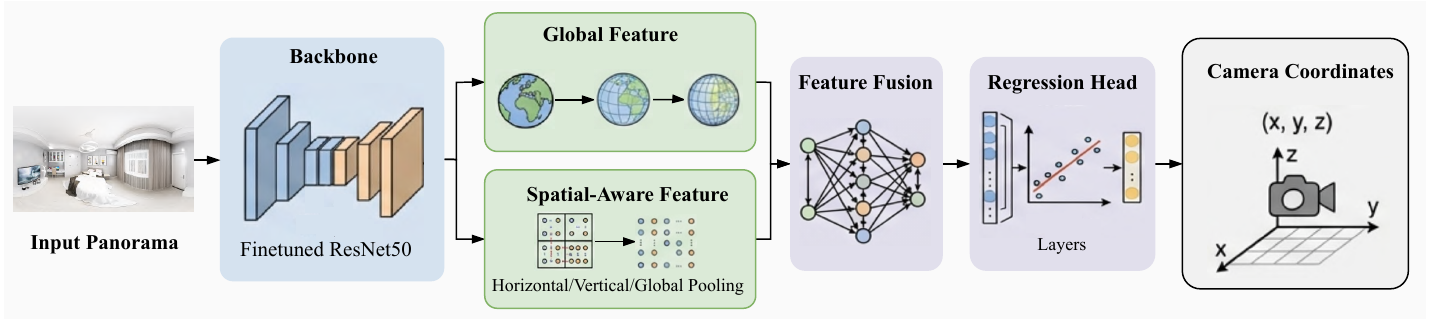}
    \caption{Architecture of CamPosNet. Given a panoramic image, a fine-tuned ResNet50 
    extracts backbone features, which are processed by a dual-branch extractor to obtain 
    global and spatial-aware features. After feature fusion, a pose 
    regression head predicts the absolute camera coordinates used to determine metric 
    scene scale.}
    \label{fig:camnet}
\end{figure}

To address this, we develop CamPosNet, a network that estimates the absolute 3D translation of the panoramic camera in the world coordinate system from the input image, supervised by ground-truth camera positions provided by Structured3D~\cite{zheng2020structured3d}. CamPosNet leverages a pre-trained ResNet50~\cite{he2016resnet} backbone with a dedicated spatial awareness enhancement module and a global context fusion module, followed by a pose regression head. Multi-level features combining spatial and semantic information are fused via multi-head self-attention before the regression head outputs the 3D camera coordinates, enabling an accurate mapping from panoramic image features to absolute scene scale, as illustrated in Fig.~\ref{fig:camnet}. Table~\ref{tab:arch_campnet} provides the detailed layer-by-layer architecture and hyperparameter settings of CamPosNet. Note that camera pose prediction in CamPosNet is used solely for absolute depth recovery and rendering scale alignment; its generalization ability beyond the training distribution 
may be limited.

\begin{table}[htbp]
    \centering
    \caption{CamPosNet Architecture Details. Each row describes one 
    functional module, including its convolutional and fully-connected 
    layer counts and key structural parameters.}
    \label{tab:arch_campnet}
    \begin{adjustbox}{max width=\textwidth}
    \small  % 由 \footnotesize 改为 \small（9pt，明确满足≥8pt要求）
    \begin{tabular}{ccccccp{6cm}}
    \toprule
    \textbf{Module} & \textbf{Kernel Size} & \textbf{Stride} & 
    \textbf{Padding} & \textbf{Activation} & \textbf{Layers} & 
    \textbf{Key Parameters \& Structure} \\
    \midrule
    ResNet50 Backbone & 7x7, 3x3 & 2, 1 & 3, 1 & ReLU & 
    \makecell[l]{Conv: 49\\FC: 0} &
    Pretrained weights IMAGENET1K\_V2; 
    final GAP and FC layers removed \\
    \midrule
    Attention Mechanism &
    \makecell[l]{1x1, 3x3\\(dil.=2), 3x3} & 1 & 0, 2, 1 & ReLU &
    \makecell[l]{Conv: 5\\FC: 1} &
    Conv: 2048$\to$512$\to$512$\to$512; 
    pooling: Horiz.(1$\times$8), Vert.(4$\times$1), 
    Global(4$\times$8); 
    feature Conv: 512$\to$256, 512$\to$128; 
    FC: (2048$\times$8 + 256$\times$8 + 128$\times$4)$\to$2048 \\
    \midrule
    Global Context & -- & -- & -- & GELU &
    \makecell[l]{Conv: 0\\FC: 2} &
    Expand (B,2048)$\to$(B,1,2048); 
    8-head MultiheadAttention (embed=2048); 
    residual + LayerNorm; 
    FFN: 2048$\to$4096$\to$2048; 
    residual + LayerNorm; 
    compress to (B,2048) \\
    \midrule
    Fusion Module & -- & -- & -- & GELU &
    \makecell[l]{Conv: 0\\FC: 1} &
    Linear(4096$\to$2048) + LayerNorm; 
    Kaiming normal init (fan\_out, relu) \\
    \midrule
    Pose Regression & -- & -- & -- & GELU &
    \makecell[l]{Conv: 0\\FC: 7} &
    Shared layer 2048$\to$1024; 
    3 branches each 1024$\to$256$\to$1; 
    Xavier uniform init, bias=0; 
    output BatchNorm1d(3, affine=True) \\
    \bottomrule
    \end{tabular}
    \end{adjustbox}
\end{table}

\subsection{End-to-End Gaussian Generation}

Many single-image 3D reconstruction methods rely on SFM-based pipelines that first synthesize multiple views and then apply COLMAP~\cite{Schonberger2016sfm} to produce 3DGS inputs. Such pipelines introduce compounding uncertainty, as scene completion depends heavily on generative model quality and consistency. In contrast, our approach directly regresses 3D Gaussian parameters from a single panoramic image via a lightweight encoder-decoder, without relying on multi-view synthesis or any test-time optimization.

\noindent\textbf{Adaptive Gauss-Generator.}
To achieve scene-adaptive control over Gaussian count and spatial distribution, we introduce the Gauss-Generator. A lightweight network estimates global image complexity $C$ (defined as the mean gradient magnitude over the full image) and local texture quantile $T_q$ (the $q$-th percentile of normalized local gradient magnitudes), which are fused to determine the target Gaussian number $G$:
\begin{equation}
    \begin{split}
        f_c &= \sigma(\lambda_c C), \quad f_t=\sigma(\lambda_t T_q),\\
        A &= \operatorname{clamp}(w_c f_c + w_t f_t,\; a_{\min},\; 1),\\
        G &= \mathrm{round}\!\big(\operatorname{clamp}(m + (M-m)A,\; m,\; M)\big),
    \end{split}
\end{equation}
where $\lambda_c, \lambda_t$ are complexity and texture scaling factors, $w_c, w_t$ are weighting coefficients ($w_c+w_t=1$), $a_{\min}$ is the lower bound, and $m, M$ define the minimum and maximum Gaussian count.

At the pixel level, a sampling density map $\rho(x)$ is constructed by combining structural and textural information:
\begin{equation}
    \rho(x) = \max\!\left(\gamma_d D(x),\;\gamma_t T(x)\right),
\end{equation}
where $T(x)$ is the normalized local gradient magnitude, $D(x)$ is the per-pixel depth normalized to $[0,1]$, and $\gamma_d, \gamma_t$ are balancing coefficients. These density values determine whether a pixel generates one or more Gaussians (via a split strategy inspired by S2Gaussian~\cite{wan2025s2gaussian}) or contributes no additional Gaussian beyond the base allocation. Normalized sampling probabilities within the valid pixel set $V$ are:
\begin{equation}
    p_i = \frac{\rho_i q_i}{\sum_{j\in V}\rho_j q_j}, \quad i \in V,
\end{equation}
where $q_i$ represents a per-pixel confidence weight derived from the depth quality map.

For pixels selected to generate multiple Gaussians (high-texture split regions), the decoder predicts a shared set of base Gaussian parameters, and additional Gaussians are obtained by small learned perturbations around the base parameters, with each pixel's additional count scaled by the global budget $G$. The decoder outputs raw parameters $(s^{r}, c^{r}, o^{r}, r^{r})$, mapped to final Gaussian attributes via:
\begin{equation}
    \begin{aligned}
        s &= \exp(\tanh(s^{r})\alpha_s)\, s_0 + \Delta_s, \quad
        c = \sigma(c^{r}),\\
        o &= \sigma(o^{r})\alpha_o+\beta_o, \quad
        r = \frac{r^{r}}{\|r^{r}\|},
    \end{aligned}
\end{equation}
where $s, c, o, r$ represent scale, color, opacity, and rotation, respectively, and $\alpha_s, s_0, \Delta_s, \alpha_o, \beta_o$ are learnable mapping parameters. This approach balances the Gaussian distribution and reduces issues of redundancy and deficiency.

\begin{figure}[t]
  \centering
  \includegraphics[width=\linewidth]{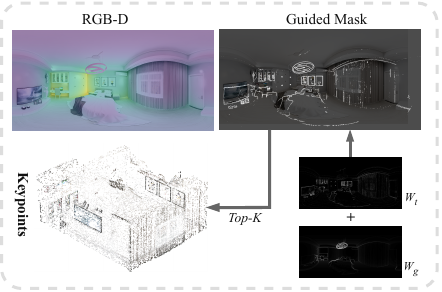}
  \caption{Candidate Keypoints. Keypoints are selected based on texture and geometric edges to provide structural constraints for the scene Gaussian distribution.}
  \label{fig:keypoints}
\end{figure}

\noindent\textbf{Point Cloud Guidance.}
To fully utilize 3D geometric information, we use Point-Guidance to enhance sampling density in key regions, as shown in Fig.~\ref{fig:keypoints}. A point cloud is obtained from the input depth map and panoramic image through spherical unprojection. The local geometric complexity is computed as the gradient magnitude over the 3D point field:
\begin{equation}
    \mathcal{G}(y, x) = \sqrt{\|\nabla_x \mathbf{P}(y, x)\|^2 + \|\nabla_y \mathbf{P}(y, x)\|^2},
\end{equation}
where $\mathbf{P}(y, x)$ is the 3D point coordinate unprojected from depth at pixel $(y,x)$, and $\nabla_x$, $\nabla_y$ denote finite-difference gradient operators in horizontal and vertical directions. High values correspond to key structures with rich geometric variation.

To jointly consider texture, edge, and geometric information, we define a pixel-level saliency map:
\begin{equation}
    S = \lambda_e \widehat{E} + \lambda_t \widehat{T} + \lambda_g \widehat{\mathcal{G}},
\end{equation}
where $\widehat{E}$, $\widehat{T}$, and $\widehat{\mathcal{G}}$ are the normalized depth edge map, texture map, and geometric complexity map, respectively, and $\lambda_e$, $\lambda_t$, $\lambda_g$ are corresponding weight coefficients summing to one. The sampling probability at each valid pixel is then:
\begin{equation}
    p_{y,x} = \frac{S(y, x)}{\sum_{(i,j) \in V} S(i, j)}, \quad (y, x) \in V,
\end{equation}
where $V$ is the set of valid (non-sky, non-invalid-depth) pixels. This importance-weighted sampling strategy reduces geometric discrepancies that arise from relying solely on 2D image features.

\noindent\textbf{Geometric Correction \& Occlusion Completion.}
Equirectangular projection introduces well-known area distortions, particularly near the poles, which can degrade depth estimation and skew the Gaussian distribution. We introduce a lightweight Geo-OccComp module to compensate for this distortion.

Given spherical coordinates $(\phi, \theta)$, where $\phi \in [-\pi/2, \pi/2]$ denotes latitude and $\theta \in [-\pi, \pi]$ denotes longitude, the area distortion factor is defined as $d_a(\phi) = \cos(\phi)$, which reflects the compression of surface area toward the poles in equirectangular projection. An equator weight $w_e(\phi) = \exp(-\phi^2 / 2\sigma^2)$ down-weights corrections near polar regions where the projection is most distorted. The depth scaling factor is then $s_d = \alpha \cdot d_a(\phi) + \beta \cdot w_e(\phi)$, and the pole penalty term is $p = \gamma \cdot (1 - \cos^2(\phi))$, which applies additional suppression at high latitudes. Together, the corrected depth is:
\begin{equation}
    D_{\mathrm{corr}} = D_0\, s_d\, (1 + p),
\end{equation}
where $D_0$ is the raw predicted depth and $\alpha, \beta, \gamma$ are learnable scalar coefficients.

For occluded areas (regions not visible from the input viewpoint), a lightweight 2D inpainting network predicts plausible depth values on the rendered cubemap faces, ensuring geometric continuity. Since our focus is on fast feed-forward reconstruction rather than generation, the quality of completion under large viewpoint shifts is limited, as noted in the limitations section.

\begin{figure}[t]
    \centering
    \includegraphics[width=\linewidth]{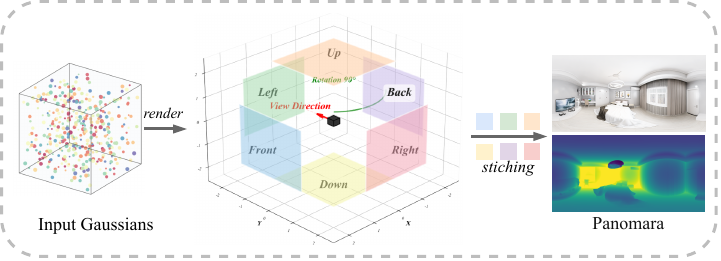}
    \caption{Cubemap rendering. The scene Gaussians are rasterized onto six cube faces 
    via fixed perspective cameras. The resulting face textures are then reprojected to 
    equirectangular format via a spherical inverse mapping to produce the final panoramic 
    render.}
    \label{fig:render}
\end{figure}

\noindent\textbf{Panoramic View Rendering.}
Since standard 3DGS~\cite{Kerbl20233dgs} rasterization assumes a perspective camera and cannot directly render panoramas, we design a CubeToERP Renderer, as shown in Fig.~\ref{fig:render}. The scene Gaussians are rasterized onto six cube faces using fixed perspective cameras oriented along the $\pm x$, $\pm y$, and $\pm z$ axes. Volume rendering produces per-pixel RGBA and depth textures for each face. An ERP-to-cube spherical map is constructed by converting ERP pixel coordinates (latitude, longitude) to 3D view vectors and projecting onto the appropriate cube face via normalized texture coordinates. The final ERP image is assembled by bilinear sampling from the six face textures and alpha blending to avoid oversaturation.

Table~\ref{tab:arch_fastpano3d} summarizes the architecture of each functional module in FastPano3D. All components are deliberately designed with lightweight building blocks—including circular padding, residual connections, and adaptive pooling—which collectively contribute to the significant inference speedup demonstrated in our experiments while keeping the total parameter count competitive with existing methods.

\begin{table}[htbp]
    \centering
    \caption{FastPano3D Module Architecture Details. All modules are 
    built with lightweight components to ensure fast feed-forward 
    inference.}
    \label{tab:arch_fastpano3d}
    \begin{adjustbox}{max width=\textwidth}
    \small  % 由 \footnotesize 改为 \small（9pt）
    \begin{tabular}{ccccccp{5cm}}
    \toprule
    \textbf{Module} & \textbf{Kernel} & \textbf{Stride} & 
    \textbf{Padding} & \textbf{Activation} & \textbf{Layers} & 
    \textbf{Key Parameters \& Structure} \\
    \midrule
    Encoder & \makecell[c]{7x7\\5x5\\3x3} & 1, 2 & 
    \makecell[c]{1, 2, 3\\circular} & 
    ReLU, Sigmoid &
    \makecell[l]{Conv: 15\\FC: 3} &
    3 residual blocks (dilation=1/2): 
    spatial + channel attention; 
    BatchNorm2d; 
    global FC: 256$\to$512$\to$128$\to$256 \\
    \midrule
    Decoder & 3x3, 1x1 & 1 & circular & 
    ReLU, Tanh, Sigmoid &
    \makecell[l]{Conv: 19\\FC: 1} &
    3 upsample blocks (scale=2); 
    AntiAliasBlock; global\_dim=256; 
    L2-norm for rotation output \\
    \midrule
    Gaussian Generation & 3x3 & 1 & circular & 
    ReLU, Sigmoid &
    \makecell[l]{Conv: 14\\FC: 3} &
    Complexity estimation, texture detection, 
    geometric density adjustment \\
    \midrule
    Point Cloud Guidance & 3x3 & 1 & circular & 
    ReLU, Sigmoid &
    Conv: 3 &
    Keypoint extraction via depth edges 
    and texture complexity \\
    \midrule
    Geometric Correction & 3x3, 1x1 & 1 & circular & 
    ReLU, Tanh, Sigmoid &
    Conv: 17 &
    feature\_dim=256; distortion encoding 
    + depth correction; pole penalty \\
    \midrule
    Occlusion Completion & 3x3 & 1 & 
    \makecell[c]{1\\circular} & 
    ReLU, Sigmoid &
    Conv: 18 &
    OccCompletionBlock (U-Net); 
    AdaptiveAvgPool2d(64,128) \\
    \bottomrule
    \end{tabular}
    \end{adjustbox}
\end{table}

\subsection{Training Objectives}
\label{sec:loss}

The full training objective is a weighted sum of image appearance losses, depth and 
geometry losses, and regularization terms:
\begin{equation}
    L_{\mathrm{total}} = L_{\mathrm{image}} + L_{\mathrm{depth}} + L_{\mathrm{other}}.
\end{equation}

\noindent\textbf{Image Losses.}
The image loss enforces appearance similarity at pixel, structural, and perceptual levels. 
We use a pixel reconstruction term (L1 with a small MSE regularizer), a multi-scale SSIM 
term, and an LPIPS perceptual term computed on longitude-wise patches of the ERP panorama:
\begin{equation}
\begin{split}
    L_{\mathrm{image}} &= w_{r}\,L_{r} + w_{s}\,L_{s} + w_{p}\,L_{p},\\
    L_{r} &= \|I_{\mathrm{pred}}-I_{\mathrm{gt}}\|_{1} 
             + \lambda_{\mathrm{mse}}\|I_{\mathrm{pred}}-I_{\mathrm{gt}}\|_{2}^{2},
\end{split}
\end{equation}
where $L_{s}=1-\mathrm{MS\text{-}SSIM}(I_{\mathrm{pred}},I_{\mathrm{gt}})$ and $L_{p}$ 
denotes the LPIPS loss averaged over longitudinal patches. $I_{\mathrm{pred}}$ and 
$I_{\mathrm{gt}}$ denote the predicted and ground-truth ERP panoramas, respectively.

\noindent\textbf{Depth and Geometry Losses.}
ERP pixel coordinates $(\phi,\theta)$ (latitude and longitude) are first mapped to unit 
directions and then to metric 3D points by
\begin{equation}
\begin{split}
    \mathbf{u}(\phi,\theta) &= \big(\cos\phi\cos\theta,\; 
                                    -\cos\phi\sin\theta,\; \sin\phi\big),\\
    \mathbf{P}(i) &= \mathbf{u}(\phi_i,\theta_i)\,d(i) + \mathbf{c},
\end{split}
\end{equation}
where $\mathbf{c}$ is the camera centre and $d(i)$ is the predicted depth at pixel $i$. 
A masked 3D L1 loss is applied to the predicted and ground-truth point clouds:
\begin{equation}
    L_{3\mathrm{D}} = \frac{\sum_i m_i\|\mathbf{P}_{\mathrm{pred}}(i)
                      -\mathbf{P}_{\mathrm{gt}}(i)\|_{1}}{\sum_i m_i + \varepsilon},
\end{equation}
where $m_i$ is an optional validity mask and $\varepsilon$ is a small constant for 
numerical stability. To mitigate ERP polar distortion, we additionally apply a 
latitude-weighted 2D depth loss with $w(\phi)=\max(\cos\phi,\delta)$:
\begin{equation}
    L_{2\mathrm{D}}^{\mathrm{dw}} = \frac{1}{N}\sum_i 
    w(\phi_i)\,|d_{\mathrm{pred}}(i)-d_{\mathrm{gt}}(i)|,
\end{equation}
where $\delta$ is a small clamping constant. These two terms are combined into the primary 
geometric loss $L_{\mathrm{geo}} = L_{3\mathrm{D}} + \alpha\,L_{2\mathrm{D}}^{\mathrm{dw}}$.

We further employ a scale-invariant log-space depth loss $L_{\mathrm{si}}$ to preserve 
relative depth structure. To suppress panorama seam artifacts, pixels near the seam 
(detected via a small longitudinal offset and cube-face comparison) are penalized by a 
Smooth-L1 loss on their relative depth discontinuity:
\begin{equation}
\begin{split}
    r(i) &= \frac{|d(i)-d_{\mathrm{neigh}}(i)|}
             {\tfrac{1}{2}(d(i)+d_{\mathrm{neigh}}(i))+\varepsilon},\\
    L_{\mathrm{seam}} &= \operatorname{mean}_{i\in\mathrm{seam}}
                         \,\mathrm{SmoothL1}\big(r(i),\,0\big).
\end{split}
\end{equation}
Pole regions are treated as seams to reduce ambiguity near the ERP top and bottom. 
Surface normals estimated from local depth derivatives are further regularized by a 
neighbor-normal consistency term $L_{\mathrm{normal}}$. The full depth and geometry 
objective is:
\begin{equation}
    L_{\mathrm{depth}} = w_{d}\big(L_{\mathrm{geo}} + \beta\,L_{\mathrm{si}}\big)
                       + w_{\mathrm{seam}}\,L_{\mathrm{seam}} 
                       + w_{n}\,L_{\mathrm{normal}}.
\end{equation}

\noindent\textbf{Gaussian Regularization.}
To stabilize Gaussian scale and opacity distributions during training, we apply two 
lightweight regularization terms. Denoting per-component scales by $s_g$ and opacities 
by $p_g$:
\begin{equation}
\begin{split}
    L_{\mathrm{s\_reg}}   &= \frac{1}{G}\sum_{g}
                                  \mathrm{mean}\!\big(\max(s_g - s_0,\,0)\big),\\
    L_{\mathrm{op\_reg}} &= -\frac{1}{G}\sum_{g}
                                  \mathrm{mean}\!\big(p_g\log p_g\big),\\
    L_{\mathrm{other}}        &= w_{\mathrm{s\_reg}}\,L_{\mathrm{s\_reg}} 
                                + w_{\mathrm{op\_reg}}\,L_{\mathrm{op\_reg}},
\end{split}
\end{equation}
where $s_0$ is a small scale threshold, $G$ is the total number of Gaussians, and all 
logarithms are computed with a clamp of $\max(\cdot,10^{-6})$ for numerical stability.

%%------------------------------------------------------------
\section{Experiment}
\label{sec:exp}

\subsection{Implementation Details}

We use the pre-trained EGFormer~\cite{Yun2023egformer} for monocular panoramic depth estimation; its weights are kept frozen throughout all training stages. FastPano3D is then applied for scene reconstruction to infer Gaussians and render the results. All experiments are conducted on a single NVIDIA A6000 GPU.
\begin{figure}[!ht]
    \centering
    \includegraphics[width=\linewidth]{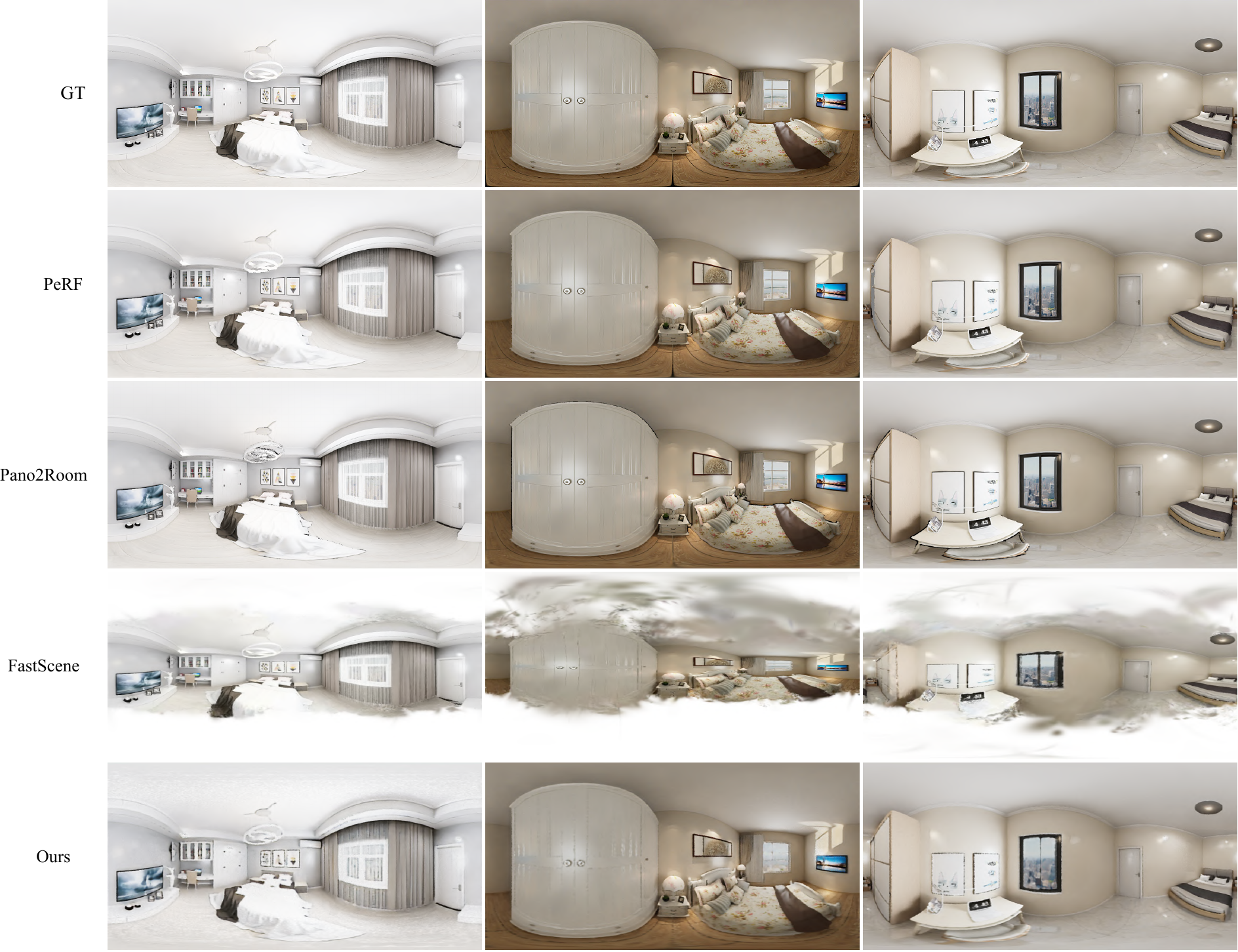}
    \caption{Qualitative panoramic comparison with other methods. For each method, we show the panoramic rendering results of three scenarios. Our method is below PERF and Pano2Room in quality (which benefit from expensive per-scene optimization), but outperforms FastScene, while being substantially faster than all compared methods.}
    \label{fig:quanjing}
\end{figure}
\noindent\textbf{Dataset.}
We use the Structured3D dataset~\cite{zheng2020structured3d} as the primary training source, which provides approximately five thousand photo-realistic indoor panoramic images with ground-truth depth maps and camera poses, split into training, validation, and test subsets. We evaluate all methods on two public benchmarks: Structured3D (in-distribution) and Replica~\cite{straub2019replica} (cross-dataset generalization). Replica serves as the standard evaluation benchmark consistently adopted across the panoramic indoor reconstruction literature, including all compared methods in this work; we follow this convention to ensure fair and direct comparison.

\noindent\textbf{Training.}
The model is trained in four progressive stages as described in Section~\ref{sec:exp}, with each stage adding one functional module. Global joint fine-tuning is applied in the final stage to stabilize inter-module interactions. Total training takes approximately 40 epochs and around 20 hours on a single NVIDIA A6000 GPU. Loss weights are determined through moderate empirical tuning rather than exhaustive grid search; the selected values remain fixed across all reported experiments.
\begin{figure}[!ht]
    \centering
    \includegraphics[width=\linewidth]{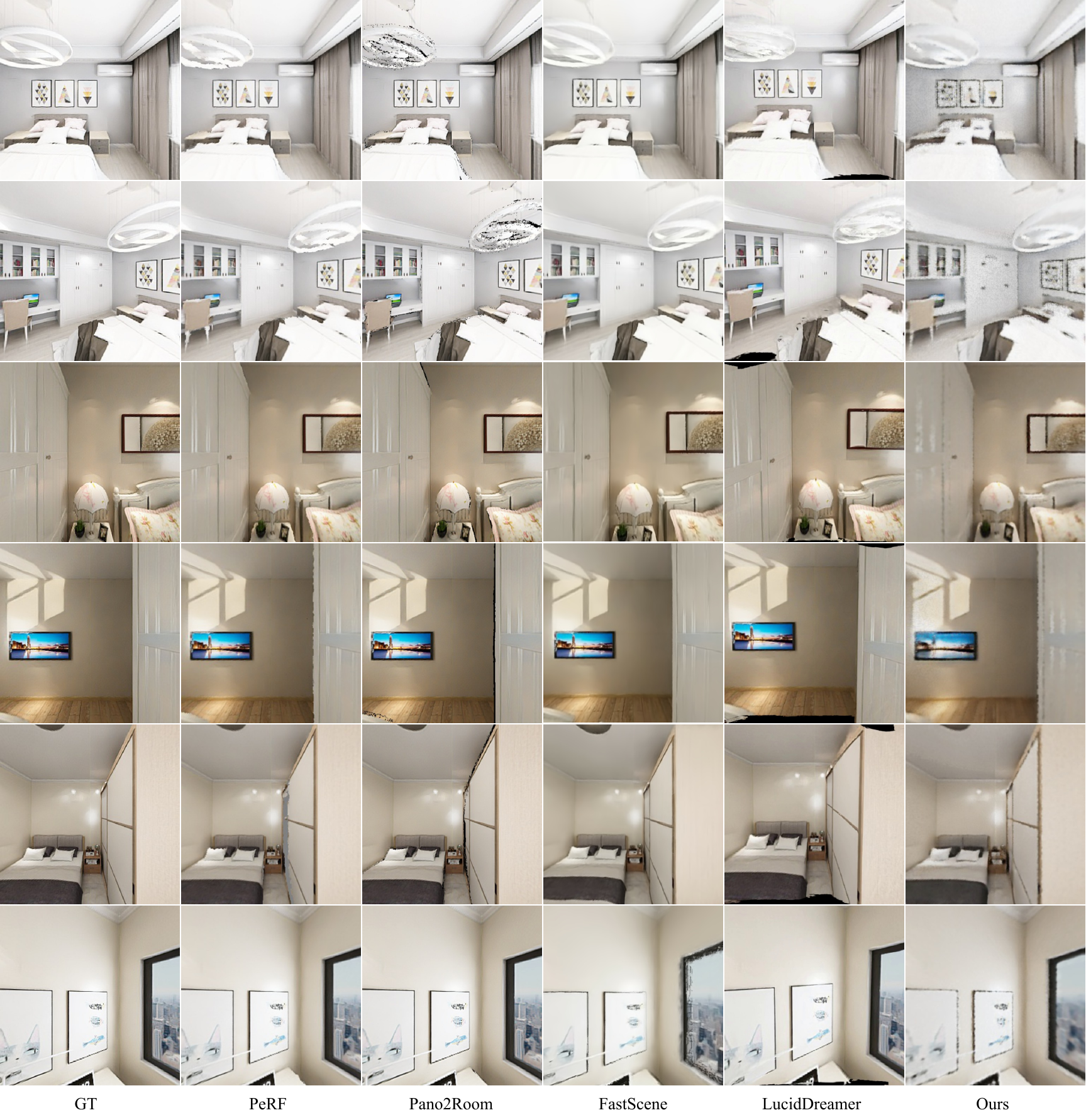}
    \caption{Qualitative perspective comparison with other methods. Consistent with the panoramic results, our visual quality is competitive with existing methods while requiring only a fraction of their reconstruction time.}
    \label{fig:toushi}
\end{figure}
\noindent\textbf{Evaluation Protocol.}
We report reconstruction quality under both panoramic and perspective views using peak signal-to-noise ratio (PSNR), structural similarity index (SSIM), and learned perceptual image patch similarity (LPIPS). For all methods, novel-view evaluation is performed with camera translations sampled from the ground-truth viewpoint distribution of each dataset. Reconstruction time is measured end-to-end from image loading to final render completion, covering all pipeline stages including depth estimation, scale prediction, Gaussian generation, and rendering. Parameter counts cover every trainable component of each method; the pre-trained EGFormer depth estimator used in our pipeline is excluded, as it functions as a fixed off-the-shelf module analogous to standard preprocessing steps in other methods.

\subsection{Comparisons with State of the Art}

We compared our method with four advanced approaches: PERF~\cite{wang2023PERF}, Pano2Room~\cite{Guo2024pano2room}, FastScene~\cite{ma2024fastscene}, and LucidDreamer~\cite{chung2023luciddreamer}. Since LucidDreamer requires perspective input, the panoramic RGBD was converted to perspective RGBD using OpenMVG.

\noindent\textbf{Quantitative Comparison.}
Table~\ref{tab:quanjing} and Table~\ref{tab:toushi} present quantitative evaluation results on Replica. In addition to image metrics, we compare reconstruction time and parameter counts. The reported time is measured end-to-end from image loading to final render completion, covering all pipeline stages including depth estimation, scale prediction, Gaussian generation, and rendering. The parameter count covers every component of each method. Because LucidDreamer does not support panoramic rendering, we compare only its perspective views. Note that LucidDreamer relies on diffusion-based generation, which produces visually plausible results but does not guarantee pixel-level alignment with the ground truth, explaining the gap between its qualitative appearance and quantitative metrics.
\begin{figure}[!ht]
    \centering
    \includegraphics[width=\linewidth]{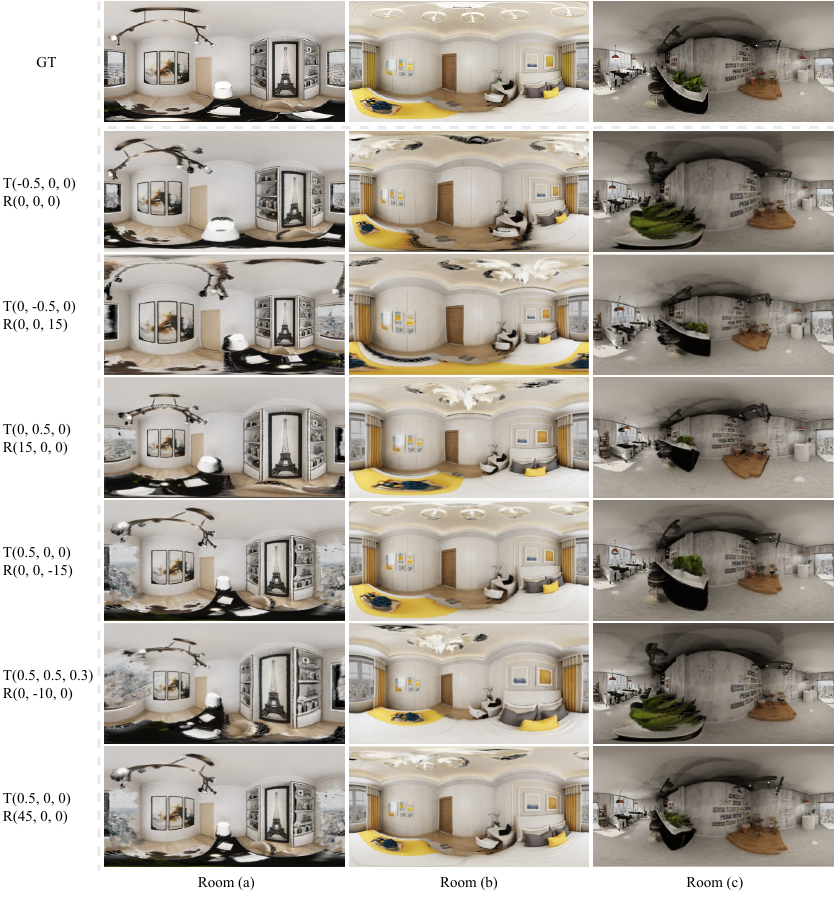}
    \caption{Novel-view rendering results across additional indoor scenes. The top row 
    presents ground-truth panoramas; subsequent rows display renders from novel viewpoints 
    obtained by applying varying degrees of translation and rotation to the original camera 
    pose. The reconstructed scene structure is qualitatively close to the ground truth, 
    and novel-view renders remain geometrically plausible across moderate viewpoint 
    changes.}
    \label{fig:novel}
\end{figure}
Our method achieves the fastest reconstruction by a wide margin, completing single-view panoramic scene inference in approximately 15 seconds, which corresponds to a \textbf{156× speed-up} over Pano2Room (2341 s) and a \textbf{212× speed-up} over PERF (3182 s). In terms of quality, our PSNR is lower than PERF and Pano2Room, which both rely on time-intensive per-scene optimization; however, our method clearly outperforms FastScene and LucidDreamer while using only half the parameters of Pano2Room. We consider this a favorable trade-off for applications that prioritize reconstruction speed. 

\begin{table}[t]
  \centering
  \caption{Quantitative panoramic comparison on the Replica dataset. Best results are highlighted in dark gray, second in medium gray, and third in light gray. LucidDreamer does not support panoramic rendering and is therefore excluded from panoramic metrics; perspective-view results are reported in Table 2.}
  \label{tab:quanjing}
  \small
  \begin{tabular}{lcccc}
    \toprule
    Method & PSNR$\uparrow$ & SSIM$\uparrow$ & LPIPS$\downarrow$ & Time (s)$\downarrow$ \\
    \midrule
    PeRF           & \cellcolor{top1}33.21 & \cellcolor{top1}0.970 & \cellcolor{top1}0.042  & 3182 \\
    Pano2Room      & \cellcolor{top2}31.51 & \cellcolor{top2}0.956 & \cellcolor{top2}0.089  & 2341 \\
    FastScene      & 11.61 & 0.572 & 0.575  & \cellcolor{top2}876  \\
    Luciddreamer   & ---   & ---   & ---    & \cellcolor{top3}2004 \\
    Ours           & \cellcolor{top3}27.96 & \cellcolor{top3}0.892 & \cellcolor{top3}0.186  & \cellcolor{top1}\textbf{15} \\
    \bottomrule
  \end{tabular}
\end{table}

\begin{table}[t]
  \centering
  \caption{Quantitative perspective comparison on the Replica dataset. P(M) denotes the total parameter count in millions. Results are consistent with the panoramic comparison. Parameter counts cover all components of each method; the pre-trained EGFormer depth estimator used in our pipeline is excluded, as it functions as a fixed off-the-shelf module analogous to standard preprocessing steps in other methods.}
  \label{tab:toushi}
  \small
  \begin{tabular}{lcccc}
    \toprule
    Method & PSNR$\uparrow$ & SSIM$\uparrow$ & LPIPS$\downarrow$ & Params (M)$\downarrow$ \\
    \midrule
    PeRF           & \cellcolor{top1}35.41 & \cellcolor{top1}0.968 & \cellcolor{top1}0.053 & 1541.82 \\
    Pano2Room      & \cellcolor{top2}33.31 & \cellcolor{top2}0.953 & \cellcolor{top2}0.078 & \cellcolor{top3}246.61  \\
    FastScene      & 24.61 & 0.825 & 0.196 & \cellcolor{top1}63.52   \\
    LucidDreamer   & 23.43 & 0.801 & 0.244 & 1066.73  \\
    Ours           & \cellcolor{top3}28.24 & \cellcolor{top3}0.905 & \cellcolor{top3}0.182 & \cellcolor{top2}132.78  \\
    \bottomrule
  \end{tabular}
\end{table}

\noindent\textbf{Qualitative Comparison.}
Fig.~\ref{fig:quanjing} and Fig.~\ref{fig:toushi} show a qualitative comparison of rendering results under panoramic and perspective views. PERF achieves the best texture and visual quality but lacks explicit 3D geometry and is the most time-consuming. Pano2Room is slightly worse in quality, showing local shadows and minor discontinuities, and is also slow. FastScene is faster but limited to perspective rendering; using 360GS~\cite{bai2024360gs} for panoramic rendering yields poor results. LucidDreamer, designed for scene generation, also cannot render panoramas and may produce small holes in perspective views. Our method is below PERF and Pano2Room in quality, which is expected given those methods spend orders of magnitude more time on per-scene optimization, but clearly outperforms FastScene and LucidDreamer, achieving competitive reconstruction quality through fully feed-forward Gaussian generation.

Fig.~\ref{fig:novel} further presents novel-view rendering results across additional indoor scenes beyond those shown in the main comparison. The top row displays ground-truth panoramas, while subsequent rows show renders obtained by applying varying degrees of translation and rotation to the original camera pose. The reconstructed structures are qualitatively close to the ground truth, and novel-view renders remain geometrically plausible across moderate viewpoint changes. Since FastPano3D prioritizes feed-forward reconstruction speed over scene completion, minor unfilled holes may appear in heavily occluded regions—a limitation shared with existing generative approaches under large viewpoint shifts. We deliberately forgo generative inpainting to reduce data dependency and preserve the core advantage of fast, single-pass inference.

\begin{figure}[!ht]
  \centering
  \includegraphics[width=\linewidth]{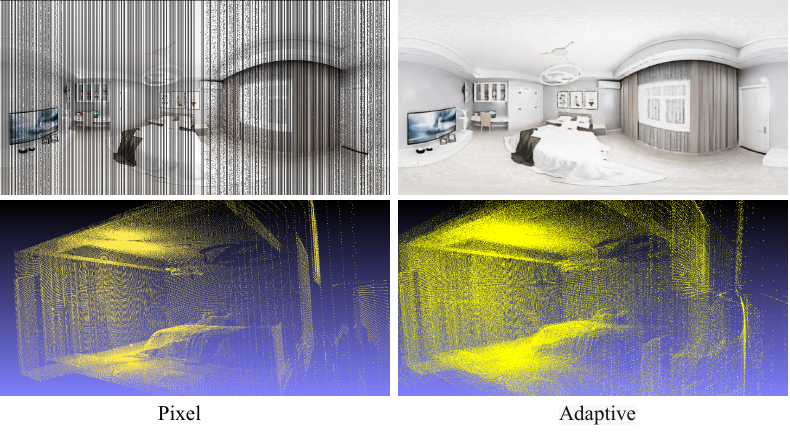}
  \caption{Comparison of Gaussian sampling strategies. Without the Gauss-Generator, per-pixel sampling causes severe discontinuities (black vertical lines). Adaptive sampling ensures scene consistency by allocating more Gaussians to texture-rich and geometrically complex regions.}
  \label{fig:gaussian}
\end{figure}

\subsection{Ablation Study}

Table~\ref{tab:component} shows the activation status of each module across training stages. Stages 1–3 progressively introduce each component, while Stage~4 shares the same module configuration as Stage~3 but applies global joint fine-tuning across all components to stabilize inter-module interactions. All ablation experiments are conducted on the Replica dataset.
\begin{table}[!t]
  \centering
  \caption{Module activation across training stages. \ding{51} indicates the corresponding module is enabled at that stage.}
  \label{tab:component}
  \small
  \begin{tabular}{cccc}
    \toprule
    Stage & Gaussian Generator & Point Guidance & Geo-OccComp \\
    \midrule
    1 & \ding{51} &            &             \\
    2 & \ding{51} & \ding{51}  &             \\
    3 & \ding{51} & \ding{51}  & \ding{51}   \\
    4 & \ding{51} & \ding{51}  & \ding{51}   \\
    \bottomrule
  \end{tabular}
\end{table}
Fig.~\ref{fig:gaussian} presents qualitative results comparing per-pixel and adaptive Gaussian sampling. Without the Gauss-Generator, per-pixel sampling causes severe discontinuities under the predicted camera scale, visible as black vertical artifacts. Adaptive sampling allocates more Gaussians to texture-rich and geometrically complex regions, ensuring scene consistency. Table~\ref{tab:stage_comparison} quantifies the contribution of each module. Each additional component provides consistent gains in all three metrics, and the global fine-tuning in Stage~4 further stabilizes the result.

\begin{table}[t]
  \centering
  \caption{Ablation studies for each module across training stages on the Structured3D dataset.}
  \label{tab:stage_comparison}
  \small
  \begin{tabular}{lccc}
    \toprule
    Stage & PSNR$\uparrow$ & SSIM$\uparrow$ & LPIPS$\downarrow$ \\
    \midrule
    1 & 25.31 & 0.812 & 0.241 \\
    2 & 26.28 & 0.834 & 0.212 \\
    3 & 26.57 & 0.857 & 0.191 \\
    4 & 26.76 & 0.868 & 0.178 \\
    \bottomrule
  \end{tabular}
\end{table}

%%------------------------------------------------------------
\section{Conclusion}
\label{sec:conclusion}

We propose \textbf{FastPano3D}, an end-to-end feed-forward framework for rapid 3D 
scene reconstruction from a single panoramic image. By combining ERP-aware adaptive 
Gaussian sampling, point-cloud-guided keypoint extraction, and distortion-aware cubemap 
rendering, FastPano3D infers a complete 3D Gaussian scene representation in seconds 
without any per-scene optimization. On the Replica benchmark, FastPano3D achieves a 
\textbf{156$\times$} speed-up over Pano2Room and a \textbf{212$\times$} speed-up over 
PERF, with competitive rendering quality (PSNR 27.96 dB, SSIM 0.892 under panoramic 
evaluation) using only half the parameters of Pano2Room.

\noindent\textbf{Strengths and Broader Impact.}
The primary strength of FastPano3D lies in its fully feed-forward design: unlike all 
prior panoramic reconstruction methods that require per-scene optimization or multi-step 
generative pipelines, our framework requires a single forward pass, making it directly 
applicable in latency-sensitive scenarios such as AR/VR preview, robotic scene 
understanding, and large-scale digital twin construction. The adaptive Gaussian sampling 
strategy addresses a key challenge specific to ERP inputs---the non-uniform spatial 
density of scene information---and the point-cloud guidance module provides a 
lightweight mechanism to incorporate 3D geometric cues without requiring multi-view 
input. These design choices are generalizable beyond our specific architecture and may 
benefit future feed-forward reconstruction methods for other projection types.

\noindent\textbf{Limitations.}
Several limitations remain. First, the quality of scene completion in heavily occluded 
regions is constrained by the lightweight nature of our inpainting module; large 
viewpoint shifts can expose unfilled areas that optimization-based methods handle through 
iterative refinement. Second, CamPosNet is supervised on Structured3D and its scale 
estimation may degrade on out-of-distribution scenes with atypical room dimensions. 
Third, our evaluation is restricted to indoor environments; the ERP distortion correction 
and Gaussian density strategy have not been validated for outdoor panoramic scenes, where 
depth ranges and scene statistics differ substantially. Finally, as a feed-forward model, 
FastPano3D does not exploit per-scene appearance cues that optimization-based methods 
leverage during test time, which contributes to the quality gap on high-frequency 
texture regions.

\noindent\textbf{Future Work.}
We identify several promising directions. Incorporating a lightweight diffusion-based 
completion module for occluded regions could close the quality gap with optimization-based 
methods while preserving feed-forward speed. Extending the framework to multi-view 
panoramic inputs---taking two or more ERP images as input---would allow cross-view stereo 
cues to reduce depth ambiguity. Adapting FastPano3D to outdoor scenes requires 
addressing sky regions and unbounded depth ranges, which are fundamentally different 
from the bounded indoor geometry assumed in the current design. Finally, integrating 
recent advances in feed-forward reconstruction for perspective images, such as improved 
Gaussian parameter prediction heads and uncertainty-aware depth estimation, could 
further improve both quality and robustness of panoramic scene inference.

% \section{Declaration of Generative AI and AI-assisted Technologies in the Writing Process}
% During the preparation of this manuscript, the authors used [Tool Name, e.g., ChatGPT] 
% for the purpose of [specific use, e.g., language polishing and grammar checking]. 
% After using this tool, the authors reviewed and edited the content as necessary and 
% take full responsibility for the content of the published article.
%%------------------------------------------------------------
\bibliographystyle{elsarticle-num}
\bibliography{ref}
\end{document}